\documentclass{ecai} 
\usepackage{latexsym}
\usepackage{amssymb}
\usepackage{amsmath}
\usepackage{amsthm}
\usepackage{booktabs}
\usepackage{enumitem}
\usepackage{graphicx}
\usepackage{color}
\usepackage{grffile}
\usepackage{multirow}
\usepackage{standalone}

\usepackage{tikz}
\usetikzlibrary{
    shapes.geometric,
    arrows.meta,
    positioning,
    fit,
    backgrounds,
    shadows,
    calc,
    angles,
    quotes,
    matrix,
    decorations.pathreplacing
}
\usepackage{pgfplots}
\pgfplotsset{compat=1.18}

\usepackage[ruled,vlined]{algorithm2e}

\usepackage{algorithm}
\usepackage{algpseudocode}
\usepackage{afterpage}

\newcommand{\BibTeX}{B\kern-.05em{\sc i\kern-.025em b}\kern-.08em\TeX}

\begin{document}

%%%%%%%%%%%%%%%%%%%%%%%%%%%%%%%%%%%%%%%%%%%%%%%%%%%%%%%%%%%%%%%%%%%%%%%%

\begin{frontmatter}

%%% Use this command to specify your submission number.
%%% In doubleblind mode, it will be printed on the first page.

\paperid{123} 

%%% Use this command to specify the title of your paper.

\title{Pareto-Informed Smart ``Predict, then Optimize'' for Multi-Objective Combinatorial Problems}

%%% Use this combinations of commands to specify all authors of your 
%%% paper. Use \fnms{} and \snm{} to indicate everyone's first names 
%%% and surname. This will help the publisher with indexing the 
%%% proceedings. Please use a reasonable approximation in case your 
%%% name does not neatly split into "first names" and "surname".
%%% Specifying your ORCID digital identifier is optional. 
%%% Use the \thanks{} command to indicate one or more corresponding 
%%% authors and their email address(es). If so desired, you can specify
%%% author contributions using the \footnote{} command.

\author[A]{\fnms{Aabhash}~\snm{Dhakal}\orcid{0009-0006-1245-6824}\thanks{Corresponding Author. Email: aabhash.dhakal@uni-wuerzburg.de}}
\author[A]{\fnms{Tim}~\snm{Lachner}\orcid{0009-0001-6982-1697}}
% \author[B,C]{\fnms{Third}~\snm{Author}\orcid{....-....-....-....}} 

\address[A]{University of Würzburg}
% \address[B]{Short Affiliation of Second Author and Third Author}
% \address[C]{Short Alternate Affiliation of Third Author}

%%% Use this environment to include an abstract of your paper.

\begin{abstract}
Many real-world planning applications, from logistics, routing to resource allocation, give rise to optimization problems that need to balance multiple, conflicting objectives. While machine learning is often used to predict the unknown parameters or costs of these problems, standard models trained on prediction accuracy can lead to suboptimal plans. Decision-Focused Learning (DFL) addresses this by integrating the downstream optimization task into the training objective. However, existing DFL methods for multi-objective problems rely on simple scalarization, which fails to capture the rich trade-offs of the Pareto Frontier. To address this gap, we propose a Pareto-aware extension to single-objective Smart ``Predict, then Optimize'' (SPO{+}) for multi-objective combinatorial optimization problems. Our loss function leverages the true Pareto Frontier, forming a target decision that aligns with a sampled preference vector, and minimizes regret against this target. Experiments on multi-objective shortest path and multi-objective knapsack problems show that our approach outperforms traditional two-stage Mean Squared Error (MSE) baselines. It also surpasses scalarization-based SPO{+} approaches. 

% This work offers a robust method for learning in multi-objective planning and decision-making contexts

% enabling better alignment between learned models and complex decision-making criteria.
\end{abstract}

\end{frontmatter}

%%%%%%%%%%%%%%%%%%%%%%%%%%%%%%%%%%%%%%%%%%%%%%%%%%%%%%%%%%%%%%%%%%%%%%%%

\section{Introduction}

Decision making in complex operational domains, such as logistics, infrastructure management, or resource allocation, frequently involves navigating complex trade-offs between multiple conflicting objectives \citep{Hauer2009, greenerfaster, Mandell1991}. For example, a logistics provider must constantly balance the speed of delivery against fuel consumption and monetary cost, where the optimal choice is rarely the same for all three objectives. For problems with a discrete solution space, Multi-Objective Combinatorial Optimization (MOCO) \citep{Ahmed2012, Jia2024} provides a formal framework for finding optimal trade-offs; however, this requires the objective coefficients (i.e. costs) to be known in advance.

A fundamental difficulty in applying automated planning to real-world domains is that these crucial parameters are often uncertain and dynamic. In a transportation network, the cost to traverse a segment of a road is not a static value, but a function of time-varying characteristics such as current weather patterns, traffic congestion, time of day or whether it is a holiday \citep{Ru2023}. This uncertainty has led to the widespread adoption of "predict, then optimize" approach \citep{Mii2020, Vanderschueren2022}. In this two-stage approach, a machine learning (ML) model first predicts the unknown cost vector, $\hat{c}$, based on observable features, $x$. Subsequently, these predictions are treated as fixed parameters which are fed into a deterministic combinatorial solver to compute a supposedly optimal plan or decision, $w(\hat{c})$.

While modular and intuitive, this sequential approach suffers from a critical weakness known as the ``prediction-decision gap''\citep{Elmachtoub2022, ETOvsIEO, Shah2024}. Standard ML approaches are typically trained to maximize predictive accuracy by minimizing a loss function like Mean Squared Error (MSE) between the predicted costs, $\hat{c}$ and the true costs, $c$. The underlying assumption is that more accurate predictions will invariably lead to better decisions. However, a substantial body of research has demonstrated that this assumption can be false \citep{Elmachtoub2022, Mandi_2024}. The reason is that not all prediction errors are equal; a model with a low overall error may make small mistakes on costs critical to the solver, leading to highly suboptimal decisions, while a model with higher error on irrelevant costs may yield a near-optimal decision. This misalignment between the learning objective and the ultimate decision-making goal can lead to demonstrably poor outcomes in practice.

To bridge this gap, Decision-Focused Learning (DFL) has emerged as an alternative \citep{Mandi_2024, Schutte2024}. Recent works, such as ``Smart Predict, then Optimize'' (SPO+), directly incorporate the structure of the downstream optimization problem into the ML model's training objective \citep{Elmachtoub2022}. Instead of minimizing prediction error, they aim to minimize decision regret i.e. the quality loss incurred by making a decision based on predicted costs compared to the optimal decision that would have been made with true costs. This allows the ML model to focus on parts of the prediction that have impact on the actual solutions.

\begin{figure*}[h]
  \centering
  % \scalebox{0.8}{
        % \input{./figures/block_diagram/pipeline}
    % }
    \includegraphics[width=0.95\textwidth]{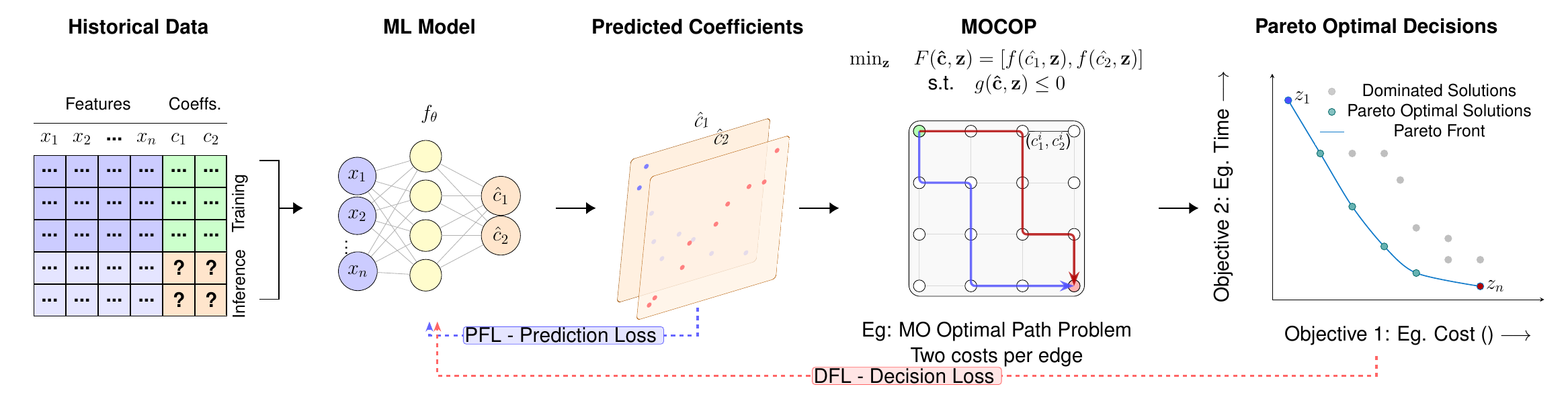}
    % \includestandalone[width=0.95\textwidth]{./conference_paper/figures/block_diagram/pipeline}
  \caption{Data Driven Multi-Objective Combinatorial Optimization - Prediction Focused Learning vs Decision Focused Learning illustrated via a bi-objective optimization example with two set of costs}
\end{figure*}

While the DFL framework has demonstrated considerable efficacy for single-objective problems its extension to multi-objective optimization contexts remains unexplored. In such problems, there is typically no single solution that is optimal across all objectives simultaneously. The concept of a single ``best'' solution is replaced by the concept of Pareto optimality. A solution is Pareto optimal if it is impossible to improve its performance on any single objective without degrading its performance on at least one other \citep{khoiadjia2013}. The set of all such solutions constitutes the Pareto set, and their corresponding values in the objective space form the Pareto Front. 

% Taking an example of the aforementioned logistics company, it is not only concerned with minimizing travel time; it must simultaneously consider fuel consumption, monetary costs, delivery reliability, and perhaps environmental impact. These objectives are often in conflict: the fastest route is rarely the most fuel-efficient, and the cheapest option may not be the most reliable. 

A common strategy for adapting single-objective DFL to a multi-objective setting is to use linear scalarization, which combines multiple objectives into one using a preference vector. This approach is fundamentally limited, as it is only capable of finding solutions on the convex hull of the Pareto Front and fails for many combinatorial problems where the front is non-convex. This leaves a significant research gap: \emph{the need for a DFL method that is both computationally tractable and can effectively handle the non-convex trade-offs inherent in multi-objective problems.}

In this paper, we introduce an extension to SPO{+} specifically designed for multi-objective combinatorial optimization problems. While there have been previous works that extend SPO+ to Multi-Task settings \citep{Tang2023} with multiple independent objectives, our work extends SPO{+} to problems with true conflicting objectives that necessitate trade-offs. We introduce a loss function that overcomes the limitations of naive scalarization in the multi objective setting, whilst maintaining its advantages. In our approach, we inform the learning process with knowledge of the true Pareto Front. An ideal target solution from the Pareto Front is selected based on cosine similarity \citep{cosmos} with a sampled Preference, whilst still solving the problem via scalarization. This provides an accurate and geometrically sound learning signal to the model, guiding it to learn cost predictions that lead to high-quality decisions even in non-convex regions of the solution space.

Our Pareto-informed approach significantly outperforms standard baselines in minimizing decision regret. We demonstrate the efficacy and generality of our approach across multiple domains, with strong performance on both the Multi-Objective Shortest Path and Multi-Objective Knapsack problems \citep{Bazgan2017}. Our implementation utilizes the popular PyEPO framework \citep{tang2024}, and extends it to generate multi-objective costs, providing a valuable baseline for future research in this area.

% The remainder of this paper is structured as follows. Section 2 reviews related work in decision-focused learning and multi-objective optimization. Section 3 formally defines the problem setting. Section 4 presents our proposed method and loss function in detail. Section 5 describes the experimental setup and presents results comparing our method against key baselines. Finally, Section 6 discusses the implications of our findings, acknowledges limitations, and concludes the paper with directions for future work.

%%%%%%%%%%%%%%%%%%%%%%%%%%%%%%%%%%%%%%%%%%%%%%%%%%%%%%%%%%%%%%%%%%%%%%%%

\section{Related Works}

% This section reviews the three core areas of research that form the foundation for our work: DFL, Multi-Objective Optimization (MOO), and Multi Objective Learning (MOL). We situate our contribution at the intersection of these domains.

DFL methods \citep{Mandi_2024}, in an effort to close the ``prediction-decision gap'', integrate downstream optimization problem directly into the model's training process, optimizing for decision quality rather than pure predictive accuracy. \citet{Mandi_2024} provide a comprehensive overview of existing DFL methods, benchmarks, challenges, and possible extensions. 

The seminal work in this area is the \textbf{Smart ``Predict, then Optimize'' (SPO+)} loss by \citet{Elmachtoub2022}.  It outperforms standard prediction-focused methods, specially under conditions of model misspecification. While the SPO loss is discontinuous and does not provide signals for gradient descent, the authors use a convex surrogate to make learning process work. They term this loss SPO{+}. In addition, the authors show that the SPO{+} is Fisher Consistent, which means, that under full distributional knowledge and mild conditions, minimizing the surrogate loss is equivalent to minimizing true SPO loss. Likewise, they also show that with increasing sample size, decision made using SPO{+} converge to full-information optimal decisions. 

\citet{Tang2023} propose an extension to Smart ``Predict, then Optimize'' for Multi-Task learning where decision error is minimized for multiple, related optimization tasks simultaneously. Their work differs from ours in that their objectives are independent. The tasks might share underlying information or predictions, but each is a different optimization problem. Multi-Task Loss focuses on training one model for many different problems, each with its own objective, while Multi-Objective Loss focuses on training a model for one problem that has many objectives within that single problem. \citet{mtmoruchte} show how Multi-Task problems are fundamentally different from Multi-objective problems.

The principles of DFL have been successfully applied to various combinatorial problems, including shortest path \citep{Elmachtoub2022}, knapsack problems \citep{Mandi2020}, and energy-cost aware scheduling \citep{Mandi2020}, demonstrating its broad applicability. \citet{tang2024} developed PyEPO, A PyTorch-based End-to-End Predict-then-Optimize Tool which provides a collection of datasets, problems and methods for solving predict-then-optimize problems with the linear objective function. We use problems and datasets from their work as inspiration for our experimental design.

% \citet{Schutte2024} present robust set of losses for DFL where they present three different strategies for making the learning process with DFL less brittle and more robust. We utilize a concept closely related of their concepts in this work to make our learning process more robust.

Separate from the DFL paradigm, there is a rich literature on Multi-Objective Machine Learning, which focuses on training models that must balance multiple, often conflicting, loss functions. A prominent area is Pareto Multi-Task Learning (MTL). \citet{lin2019pareto} propose decomposing the multi-objective problem into a set of constrained subproblems, each guided by a different preference vector. By solving these subproblems in parallel, their method can generate a set of diverse Pareto solutions. This approach of dedicating a separate optimization process for each preference vector is a direct inspiration for our work, where we train one specialized model for each known preference.

Likewise, \citet{cosmos} propose COSMOS, a method that generates the entire Pareto Front from a single model in one training run. COSMOS conditions the model by augmenting the input features with a preference vector. To ensure a well-spread Pareto Front when using linear scalarization, they introduce a penalty term that maximizes the cosine similarity between the preference vector and the resulting vector of losses. The use of cosine similarity as a mechanism for aligning a solution with a preference vector was a key inspiration for our work. However, we adapt this concept for a different purpose: rather than using it as a penalty in the loss function, we use it as a heuristic to select the most relevant target solution from the true Pareto Front to guide the SPO+ gradient.

\section{Problem Description and Proposed Approach}

Consider a supervised learning setting for a Multi-Objective Combinatorial Optimization Problem (MOCOP). For each instance $n$ in our data, we have a feature vector $x \in \mathcal{X}$. The goal is to predict a multi-objective cost vector $\hat{c} \in \mathbb{R}^{m \times K}$, where $m$ is the number of decision variables (e.g., edges in a network, or selected items in the knapsack problem) and $K$ is the number of conflicting objectives. The corresponding true cost vector is denoted by $c$. These costs are used in a combinatorial optimization problem defined on a set of feasible solutions $\mathcal{W}$. A solution $w \in \mathcal{W}$ is a vector representing a specific choice (e.g., a path in the shortest path problem). The quality of a solution $w$ is evaluated by its total cost for each objective, given by the vector $z = c^T w$. The goal of MOCOP is to find the set of non-dominated solutions, known as the Pareto set $W^* \subseteq \mathcal{W}$, and its corresponding image in the objective space, the Pareto Front $Z^*$.

\subsection{Smart ``Predict, then Optimize'' (SPO{+})}

The SPO{+} loss function is a continuous and convex surrogate for the true decision regret, making it tractable for gradient-based optimization. 

\begin{equation}
\mathcal{L}_{SPO+}(\hat{c}, c) = \max_{w \in \mathcal{W}} \{ (2\hat{c} - c)^T w - (2 \hat{c} - c)^T w^*\}
\end{equation}
% The gradient of the SPO+ loss with respect to the prediction $\hat{c}$ is given by 
% \begin{equation}
%  2(w^*(c) - w^*(2\hat{c} - c))   
% \end{equation}

The gradient of SPO{+} loss provides a powerful and direct learning signal. Geometrically, it represents a vector pointing from the suboptimal solution induced by the current prediction toward the true optimal solution, $w^*$. By backpropagating the gradient, the learning algorithm adjusts the model's parameters in a direction that causes its future predictions to induce plans that are ``closer'' to the true optimum. This is a more informative signal than that provided by MSE, which only encourages the cost vector $\hat{c}$ to be numerically closer to $c$, without regard for which components of the cost vector are actually critical for the final decision. The SPO+ loss thus focuses the model's learning capacity on correcting prediction errors that have the largest impact on decision quality.

\subsection{Multi Objective Optimization}

Multi-objective optimization deals with problems that trade-off conflicting goals. The goal is not to find a single optimal solution, but rather the set of optimal trade-offs. This is formalized by the concepts of Pareto dominance and optimality \citep{Voorneveld2003}. A solution $w_a$ is said to \emph{strictly dominate} another solution $w_b$, denoted $w_a \succ w_b$, if it is strictly better in all objectives. A solution $w_a$ \emph{dominates} $w_b$ if it is no worse than $w_b$ in all objectives and strictly better in at least one. A solution is \textbf{Pareto optimal} if no other feasible solution dominates it. The set of all such solutions forms the Pareto set, and their corresponding values in the objective space constitute the \textbf{Pareto Front} \citep{Voorneveld2003, Bazgan2017}. Predominant strategies exist for handling multiple objectives include:

\begin{enumerate}
    \item \textbf{Scalarization-Based Methods:} Scalarization approaches convert the multi-objective problem into a single-objective problem. The most common technique is Linear Scalarization (LS), also termed the weighted sum method, where a user-defined preference vector $p \in \mathbb{R}^K_+$ assigns a weight to each of the $k$ objectives, $\mathcal{L}_k(\theta)$. The problem is then reformulated as minimizing a single, scalar objective, $\min_{\theta} \sum_{k=1}^{K} p_k \mathcal{L}_k(\theta)$.
    LS is computationally efficient and allows direct application of single-objective solvers and DFL frameworks like SPO+. However, it can only find solutions on the \textit{convex hull} of the Pareto Front \citep{Jahn1985}. It is guaranteed to fail for any problems where the true Front is non-convex, which is common in combinatorial optimization.

    \item \textbf{Pareto-Based Methods:} In contrast, Pareto-based methods, such as Multi-Objective Evolutionary Algorithms (MOEAs) work by maintaining a population of solutions and using Pareto dominance as the selection criterion \citep{Deb2002}. These methods are designed to approximate the entire Pareto Front, including non-convex regions. However, they are often more computationally intensive and are not easily integrated into end-to-end, gradient-based learning frameworks. While these methods are more powerful and complete, they are often more complex and computationally intensive than aggregation-based techniques. Furthermore, their population-based, iterative nature does not lend itself as easily to integration within a gradient-based learning framework like DFL, which typically relies on a differentiable or sub-differentiable path from prediction to decision.
\end{enumerate}

This work seeks to bridge the gap between these two approaches by using a tractable, scalarization-based DFL training loop that is informed by knowledge of the complete, true Pareto Frontier. We avoid the theoretical pitfalls of naïve scalarization by using a Pareto-informed loss function, ParetoSPO loss, which selects a target from the true Pareto Front leveraging a Multi-Objective solver. This allows it to retain the computational efficiency of single-objective DFL while benefiting from the global perspective of Pareto-based methods, representing a new and effective approach to Pareto-informed, decision-focused learning. The unique nature of DFL, which has optimal decisions based on true costs, enables us to get a signal. 

\subsection{Proposed method: ParetoSPO}

% Our proposed \texttt{ParetoSPO} framework is designed for operational contexts where a known, discrete set of user preferences exists. For each of the $N$ known preferences $\{p_1, \dots, p_N\}$, we adopt a model committee architecture by training a separate, specialized model $f_{\theta_p}$. At inference time, when a user selects a policy corresponding to preference $p_i$, the dedicated expert model $f_{\theta_i}$ is invoked to predict the costs and find the optimal solution. The core innovation of our method lies in how each expert is trained, which we detail next.

% To select a single, actionable solution from the Pareto front, a decision-maker's priorities are captured by a user preference vector $p \in \mathbb{R}^K_+$. Our overall goal is to train a committee of models, $\{f_{\theta_1}, \dots, f_{\theta_N}\}$, where each model $f_{\theta_p}$ is specialized for a known preference $p$. The model $f_{\theta_p}(x)$ takes features $x$ as input to generate a cost prediction $\hat{c}$ that minimizes decision regret for its specific preference.

\begin{figure*}[h]
  \centering
  % \includestandalone[width=1\textwidth]{./conference_paper/figures/arch_new}
    \includegraphics[width=0.78\textwidth]{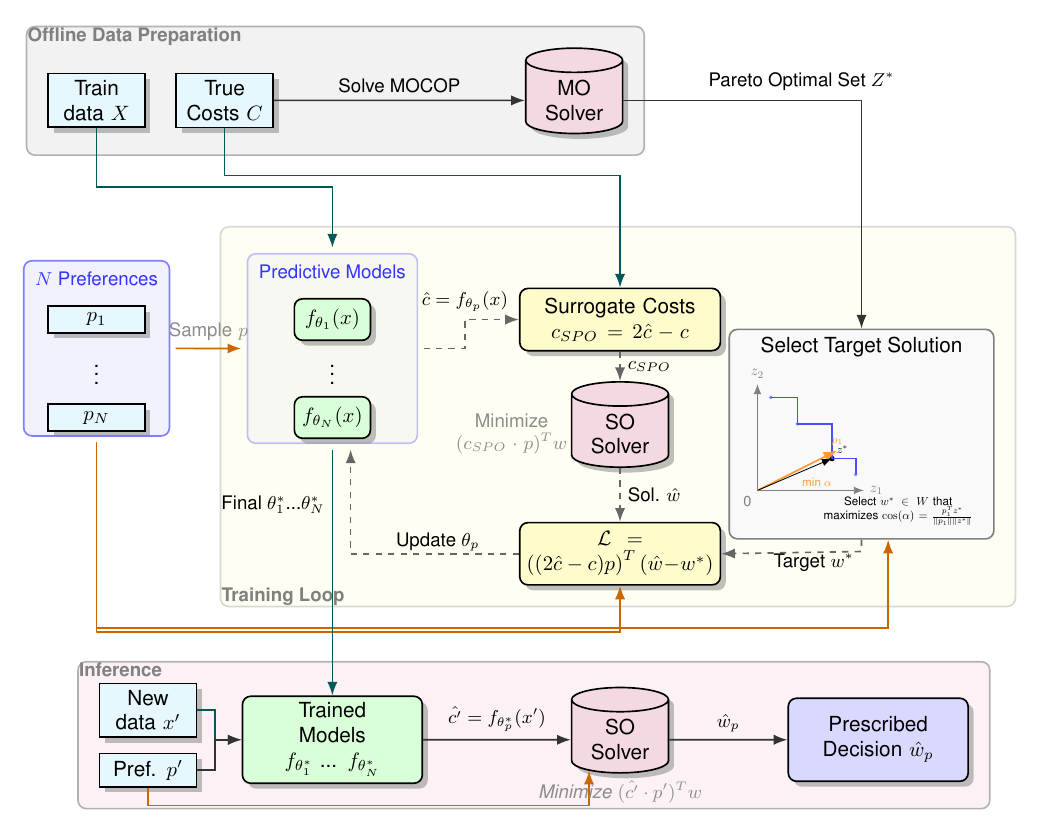}
  \caption{Pareto-informed SPO+ for Multi-Objective Decision Focused Learning}
  \label{fig:arch_new}
\end{figure*}

Our proposed \texttt{ParetoSPO} method, as outlined in Figure \ref{fig:arch_new}, is designed for operational contexts where a known, discrete set of user preferences exists. To cater to these, we train a set of models, a model committee, $\{f_{\theta_1}, \dots, f_{\theta_N}\}$, where each model $f_{\theta_p}$ is specialized for a known preference $p\in \{p_1, \dots, p_N\}$ from the set of the $N$ known preferences. This model takes features $x$ as input to generate a cost prediction $\hat{c}$ that minimizes decision regret specifically for its assigned preference $p$. 

In order to obtain the true Pareto Front for training, we use a Multi-Objective Combinatorial solver to find optimal decisions based on historical costs. Because this solver is slower than a typical single-objective one, we perform this entire calculation offline during data preparation.

During training, we select a single, actionable solution from the Pareto Front using the model's associated preference vector $p \in \mathbb{R}^K_+$. At inference time, the expert model corresponding to a given user preference is used to predict costs. 

The core innovation of our method lies in how each expert is trained, the loss function, \texttt{ParetoSPOLoss}, leverages knowledge of the true Pareto Front to guide the learning process. To understand its contribution, we first consider the standard approach: A baseline SPO+ model can be extended to handle multiple objectives via linear scalarization. In this approach, the target solution, $w^*_{scalar}$, would be defined by first scalarizing the \emph{true costs} ($c_{scalar} = c^T p$), then solving the single-objective problem. However, as established, if the true Pareto Front is non-convex, this scalarized solution may be a poor proxy for the user's desired trade-off.

\texttt{ParetoSPO} bypasses this limitation. Instead of deriving a target from scalarized costs, we select a target directly from the pre-computed true Pareto set $W^*$ and its objective values $Z^*$. The process, described in detail in Algorithm \ref{alg:paretospo}, can be summed up as follows: 

\begin{enumerate}
    \item \textbf{Predict Costs:} Given features $x$ and preference $p$, the expert model $f_\theta$ predicts the multi-objective cost vector $\hat{c}$.
    \item \textbf{Select Target Solution:} To select the ``best'' solution in the true Pareto set that aligns with the preference vector $p$, we first normalize the true objective vectors $z^* \in Z^*$ to ensure a fair comparison across objectives. Then, we compute the cosine similarity between the preference vector $p$ and each normalized true objective vector $z'^*$, obtaining solution $w^* \in W^*$, demonstrating the highest cosine similarity.

    % This heuristic finds the point on the true Pareto Front most closely aligned with the direction of the user's preference.

    \item \textbf{Compute Loss:} With the deliberately selected target $w^*$ in hand, we form a surrogate cost vector $c_{SPO} = 2\hat{c} - c$, scalarize it using the preference $p$, and solve the single-objective problem to get the predicted solution $\hat{w}$. The final loss is: $\mathcal{L} = \left((\hat{w}^T(2\hat{c} - c)  - {w^*}^T(2 \hat{c} - c)) p\right)$, shortened to $\left( (2\hat{c}-c)p \right)^T (\hat{w}-w^*)$.
\end{enumerate}

This hybrid design achieves the best of both worlds: it retains the computational efficiency of using a single-objective solver in the training loop, while transcending the theoretical limitations of simple scalarization. By providing the SPO+ gradient ($2(w^* - \hat{w})$) with a high-quality, Pareto-optimal target $w^*$, we ensure the learning signal is always directed towards a relevant region of the solution space, even if that region lies on a non-convex part of the Pareto Front. While the model committee architecture means that training time and parameter count scale linearly with the number of preferences, the impact can be mitigated by the fact that the approach is fully parallelizable.

\begin{algorithm}[H]
\caption{Pareto-Informed Decision-Focused Learning}
\label{alg:paretospo}
\KwIn{Training data $\{(x_i, c_i)\}_{i=1}^n$, preference set $\mathcal{P}$, learning rate $\eta$, number of epochs $E$}
\KwOut{Trained models $f_{\theta^*_p}$}

\vspace{0.5em}
\textbf{Offline Data Preparation:}\\
\For{$i = 1$ \KwTo $n$}{
    Compute Pareto Optimal Set: $({Z}^*_i, {W}^*_i) \gets \text{MO-Solver}(x_i, c_i)$
}

\vspace{0.5em}
\textbf{Training Loop:}\\
\For{$\text{epoch} = 1$ \KwTo $E$}{
    \For{$i = 1$ \KwTo $n$}{
        Sample preference $p \sim \mathcal{P}$\;\\
        $\hat{c} \gets f_{\theta_p}(x_i)$ \\
        $c_{\text{SPO}} \gets 2\hat{c} - c_i$ \\
        $\hat{w} \gets \arg\min_{w \in \mathcal{W}} (c_{\text{SPO}} p)^T w$ \\
        $z^* \gets \arg\max_{z \in {Z}^*} \frac{p^\top z}{\|p\| \|z\|}$,
        $w^* \gets \text{solution w} \in {W}^*, \text{corresponding to } z^*$ \\
        $\mathcal{L} \gets \left( (2\hat{c}-c)p \right)^T (\hat{w}-w^*)$ \\
        Update $\theta_p$ using $\nabla_{\theta_p} \mathcal{L}$ \;
        
    }
}
\Return{$f_{\theta^{*}_{1}}, \cdots, f_{\theta^{*}_{N}}$}
\end{algorithm}

% This hybrid design of utilizing scalarization but informing the learning process via true Multi Objective solutions retains the computational efficiency, whilst advancing performance. By providing the SPO+ gradient ($2(w^* - \hat{w})$) with a high-quality, Pareto-optimal target $w^*$, we ensure the learning signal is always directed towards a relevant region of the solution space, even if that region lies on a non-convex part of the front. This allows our method to transcend the theoretical limitations of simple linear scalarization.

% In practical terms, for a single preference, our method has no additional overhead compared to standard single objective SPO{+}. There is an argument to be made here that the need to train the model comittee, each for one preference, means the time and model parameters increase linearly with each additional preference. However, the overall architecture lends itself to parallelization, which can be used as a way to get around this.

\begin{figure*}[t]
  \centering
\includegraphics[width=1\textwidth]{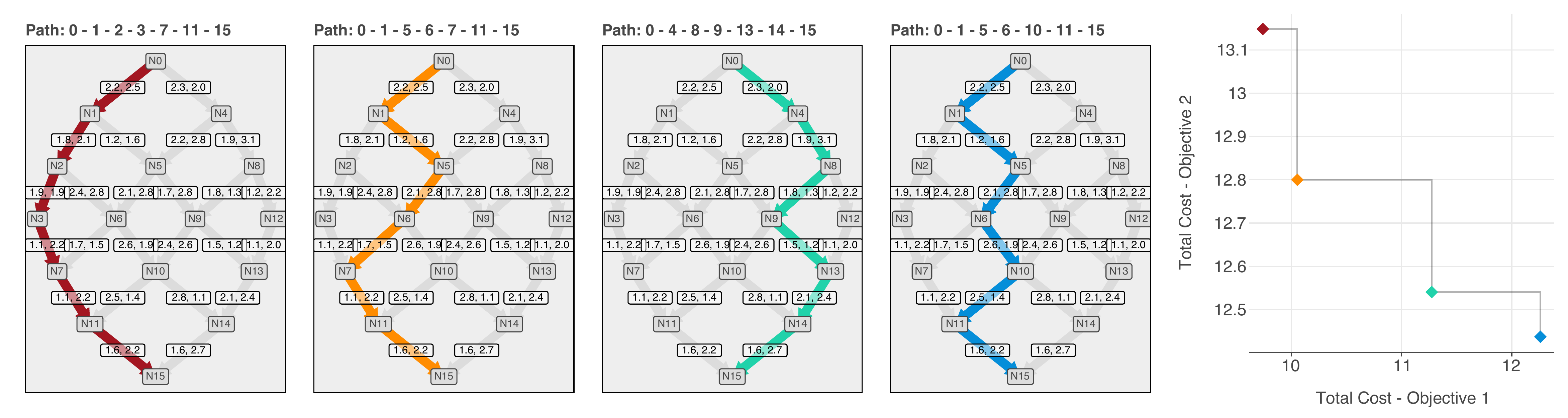}

  \caption{Example - MOSPP solved using a multi-objective solver for a 4x4 grid. Each edge in the network has two costs $(c_1, c_2)$ associated with it. A Multi Objective solver produces a set of Pareto Optimal solutions, each offering a different trade-off of total costs. Path 0-1-2-3-7-11-15 (red) offers the cost that minimizes total cost 1, the first objective, while Path 0-1-5-6-10-11-15 (blue) minimizes the second objective. The other two solutions lie in between.}
\label{fig:op-mo}
\end{figure*}

\afterpage{\begin{figure*}[h]
  \centering
  \includegraphics[width=1\textwidth]{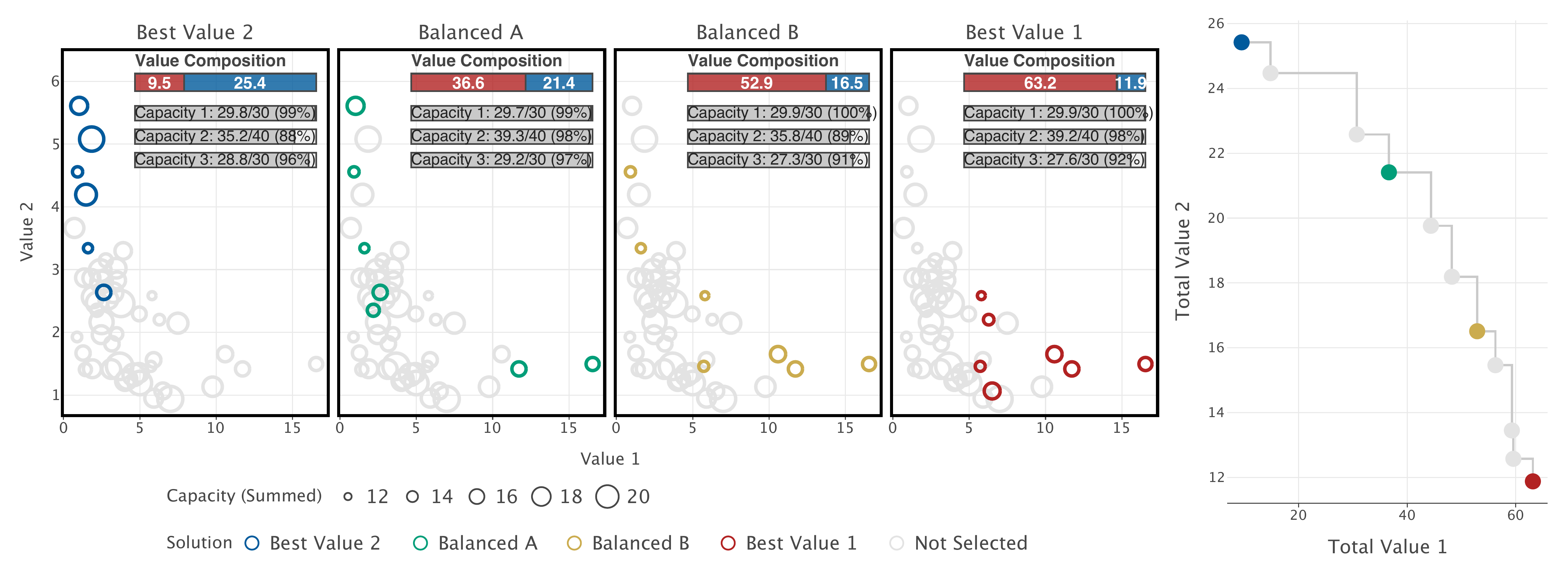}
  \caption{Example - MOMDKP problem solved using a multi-objective solver for fixed coefficients (values). Here, each item has two values associated with it, eg. monetary value, sentimental value, and 3 constraints with capacities of [30, 40, 30], eg. weight, volume, quantity, of the item. A Multi Objective knapsack solver returns a set of Pareto Optimal solutions with different tradeoffs.}
  \label{fig:knapsack-mo}
\end{figure*}}

\section{Experimental Setup}

To evaluate the performance of our proposed \texttt{ParetoSPO} method, we conduct experiments on two classic multi-objective combinatorial optimization problems: the Multi-Objective Shortest Path Problem (MOSPP) and the Multi-Objective Multi-Dimensional Knapsack Problem (MOMDKP). These problems are chosen for their relevance to real-world applications and their well-understood structures.

\subsection{Training Configuration}

We implement a two-layer feedforward network with 64 hidden units. For both problems, we use a dataset of 2000 instances, split into 70\% for training, 10\% for validation, and 20\% for testing. The models are trained for up to 100 epochs; we select the checkpoint with the best validation performance and report its results on the test set. 
To solve the underlying multi-objective combinatorial optimization problems, we employ the CP-SAT solver from OR-Tools \citep{cpsatlp_cpaior_masterclass}. For the MOSPP problem, we use a grid size=(5x5) and, for the MOMDKP, we use nitems=32. For the single-objective solver in the shortest path problem, we use a more efficient Bellman-Ford algorithm \citep{Bellman1958} in place of a generic LP solver.

\subsection{Baselines}

For both problems, we compare against three baselines:
\begin{enumerate}
    \item \textbf{Average Cost Baseline:} A non-learning baseline that uses the historical mean of objective coefficients to derive decisions. An improvement over this baseline confirms that feature-based learning is effective. 
    \item \textbf{Two-Stage (MSE):} A standard ``predict-then-optimize'' approach. First, a model is trained to predict the objective vectors by minimizing Mean Squared Error (MSE). Second, these predicted objectives are passed to the solver.
    \item \textbf{SPO+ (LS):} A decision-focused baseline that applies the standard SPO+ loss to a linearly scalarized version of the multi-objective problem. This represents the most direct adaptation of existing DFL methods to the multi-objective setting.
\end{enumerate}

\subsection{Multi-Objective Combinatorial Problems}

\subsubsection{Multi-Objective Shortest Path Problem (MOSPP)}

The MOSPP is a fundamental problem in network optimization. Given a directed graph $G=(V, E)$, the goal is to find a path from a start node $s \in V$ to a terminal node $t \in V$ that simultaneously minimizes a set of $k$ objectives. Each edge $e \in E$ has a cost vector $c_e \in \mathbb{R}^k_+$. A solution is a binary vector $w \in \{0, 1\}^{|E|}$ indicating the selected edges in a path. The problem can be formulated as:
\begin{equation}
\begin{aligned}
& \underset{w}{\text{minimize}}
& & \left( \sum_{e \in E} c_{e,1} w_e, \dots, \sum_{e \in E} c_{e,k} w_e \right) \\
& \text{subject to}
& & \sum_{e \in \delta^+(v)} w_e - \sum_{e \in \delta^-(v)} w_e = \begin{cases} 1 & \text{if } v=s \\ -1 & \text{if } v=t \\ 0 & \text{otherwise} \end{cases}, \quad \forall v \in V \\
& & & w_e \in \{0, 1\}, \quad \forall e \in E
\end{aligned}
\end{equation}
where $\delta^+(v)$ and $\delta^-(v)$ are the sets of outgoing and incoming edges for node $v$, respectively. For our experiments, we use $k=2$ objectives on a grid graph. An example of a solved MOSPP problem with known costs is shown in Figure \ref{fig:op-mo}, alongside the corresponding Pareto Frontier.

\subsubsection{Multi-Objective Multi-Dimensional Knapsack Problem (MOMDKP)}
The MOMDKP is a generalization of the classic knapsack problem and serves as a benchmark for resource allocation. Given a set of $m$ items, the goal is to select a subset to pack into a knapsack with $q$ capacity constraints, while maximizing $k$ value objectives. A solution $w \in \{0, 1\}^m$ indicates which items are selected. Formally, the problem is:
\begin{equation}
\begin{aligned}
& \underset{w}{\text{maximize}}
& & (V_1^T w, \dots, V_k^T w) \\
& \text{subject to}
& & C_j^T w \leq K_j, \quad \forall j \in \{1, \dots, q\} \\
& & & w_i \in \{0, 1\}, \quad \forall i \in \{1, \dots, m\}
\end{aligned}
\end{equation}
 where $V_1, \dots, V_k \in \mathbb{R}^m_+$ are the value vectors, $C_j \in \mathbb{R}^m_+$ are the weight vectors, and $K_j$ are the capacities. To fit this into our cost-minimization method, we treat the values as negative costs, i.e., $c = -V$. We use $k=2$ objectives and $q=2$ constraints. An example of a solved MOMDKP problem with known costs is shown in Figure \ref{fig:knapsack-mo}, alongside the corresponding Pareto Frontier.

\subsection{Data Generation}

We use datasets from the PyEPO framework \citep{tang2024} as a foundation for both problems. We extend these datasets by generating two-dimensional coefficients from external features, applying a similar method to produce costs for the MOSPP and values for the MOMDKP.

% We use the PyEPO framework \citep{tang2024} for generating the first set of coefficients, and use a similar idea to extend to a second dimension. 

For $n$ instances, we generate a feature vector $x \in \mathbb{R}^d$ from a standard normal distribution, $x \sim \mathcal{N}(0, I_d)$, where $d$ is the number of feature dimensions. A shared underlying factor, $Bx$, is computed via a linear transformation, $Bx = \frac{x B^T}{\sqrt{d}} \in \mathbb{R}^{\nu } $, where $B \in \{0, 1\}^{\nu  \times d}$ is a fixed binary matrix from a Bernoulli(0.5) distribution, and $\nu$ is the number of decision variables ($\nu =|E|$ for MOSPP, $\nu =m$ for MOMDKP). Two conflicting coefficient vectors(cost/value), $c_1$ and $c_2$, are then defined as opposing functions of this factor, controlled by a degree parameter, $\text{deg}$:
\begin{align}
c_1 = \left[ \alpha_1 \frac{(Bx + \beta_1)^{\text{deg}}}{\gamma_1^{\text{deg}}} + 1.0 \right] \odot \zeta_1,\\
c_2 = \left[ \alpha_2 \frac{(\beta_2 - Bx)^{\text{deg}}}{\gamma_2^{\text{deg}}} + 1.0 \right] \odot \zeta_2
\end{align}

% The specific scaling parameters $(\alpha, \beta, \gamma)$ differ slightly between the two problems to maintain consistency with the PyEPO framework, but the fundamental conflicting structure is the same. 
The scaling parameters $(\alpha, \beta, \gamma)$ are consistent for both the MOSPP and MOMDKP problems, with the sole exception of $\alpha_1$, which is set to 1 and 5 respectively to maintain consistency with the PyEPO framework. The shared parameter values are $(\alpha_2=1, \beta_1=3, \beta_2=5, \gamma_1=3.5, \gamma_2=5)$. Terms $\zeta_1, \zeta_2$ represent multiplicative noise vectors that simulate real-world stochasticity, where each element is drawn independently from a uniform distribution $\zeta \sim U(1-\epsilon, 1+\epsilon)$, with $\epsilon$ being the noise width. In the next section, we show our results across different choices of $\epsilon$.

\subsection{Performance Metric: Hypervolume Regret}

\begin{figure}
    \centering
    % \vspace{-15pt} % Adjust vertical spacing to pull the figure up
    \includegraphics[width=0.5\linewidth]{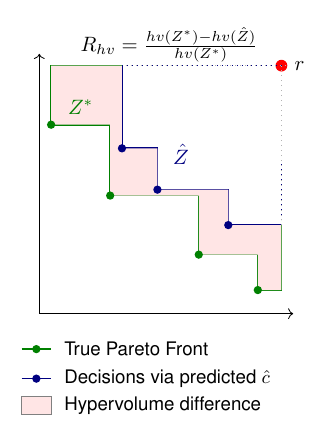}
    % \includestandalone[width=0.5\linewidth]{/regret_hv}
    \caption{Visual representation of the Hypervolume Regret.}
    \label{fig:regret-hv}
\end{figure}

We evaluate the performance of our approach using Hypervolume Regret. The Hypervolume (HV) indicator \citep{hvind} is a widely used metric for evaluating the quality of a set of Pareto Front points. The hypervolume $hv(Z)$, measures the volume of the objective space dominated by the points in a set $Z$ and bounded by a reference point $r$.
Given the true Pareto Front in the objective space, $Z^*$, and the set of decisions made with predicted costs, $\hat{Z}$, we define the hypervolume regret $R_{hv}$ as the percentage of hypervolume lost relative to the true hypervolume:

\begin{equation}
    R_{hv} = \frac{hv(Z^*) - hv(\hat{Z})}{hv(Z^*)}
\end{equation}
where $hv(Z) = \text{volume}\left(\bigcup_{z \in Z} \{q \in \mathbb{R}^2 | z \succeq q \succeq r\}\right)$.
This normalized metric is highly interpretable: a value of 0 indicates a perfect prediction of the Pareto Front, while a value of 1 indicates that the set of decisions is unable to cover any proportion of the trade-off space. Figure \ref{fig:regret-hv} illustrates the concept of the hypervolume regret for a minimization problem. For maximization problems, the objective space is conceptually inverted, and the hypervolume is calculated with respect to a nadir point (e.g., the origin) instead of a utopian reference point.

% Figure \ref{fig:regret-hv} provides a visual depiction of this metric.

\section{Results}

We present the results of our experiments in Table \ref{tab:results}. The performance of each method is evaluated based on the mean Hypervolume Regret ($R_{hv}$) across the test set, under varying levels of noise.

\begin{table}[h]
  \centering
  \caption{Mean Hypervolume Regret (\%) for MOSPP and MOMDKP. We report the test data performance with each model finalized based on validation set performance. Lower value is better; a value of 0 implies model recovered the optimal Pareto Frontier.}
  \label{tab:results}
  \begin{tabular}{@{}llccc@{}}
    \toprule
    \textbf{Problem} & \textbf{Method} & \textbf{$\epsilon$ = 0} & \textbf{$\epsilon$ = 0.25} & \textbf{$\epsilon$ = 0.5} \\
    \midrule
    \multirow{4}{*}{MOSPP} & Avg. Cost &  13.90 $\pm$ 2.69 & 17.28 $\pm$ 3.18 & 24.57 $\pm$ 4.14\\
                           & MSE & 1.15 $\pm$ 0.06 & 5.51 $\pm$ 0.23 & 13.43 $\pm$ 0.56\\
                           & SPO+ (LS) & 0.98 $\pm$ 0.07 & 5.10 $\pm$ 0.26 & 12.47 $\pm$ 0.48\\
                           & \textbf{ParetoSPO} & \textbf{0.70 $\pm$ 0.04} & \textbf{4.82 $\pm$ 0.20} & \textbf{12.00 $\pm$ 0.31} \\
    \midrule
    \multirow{4}{*}{MOMDKP} & Avg. Cost & 32.87 $\pm$ 6.46 & 37.53 $\pm$ 7.76 & 46.63 $\pm$ 8.15 \\
                           & MSE & 12.65 $\pm$ 4.60 & 20.41 $\pm$ 6.35 & 33.48 $\pm$ 5.96\\
                            & SPO+ (LS) & 10.96 $\pm$ 4.20 & 19.27 $\pm$ 6.03 & 31.80 $\pm$ 5.96\\
                            & \textbf{ParetoSPO} & \textbf{8.36 $\pm$ 3.32} & \textbf{16.67 $\pm$ 5.10} & \textbf{30.98 $\pm$ 5.19} \\
    \bottomrule
  \end{tabular}
\end{table}

% \begin{figure}
%   \centering
%   \includegraphics[width=\linewidth, height=0.3\textheight]{conference_paper/figures/results/hv_regret_plot_final_color.pdf}
%   \caption{Hypervolume Regret comparison on MOSPP and MOMDKP, we run each method for 8 random seeds on 3 different noise settings [0, 0.25, 0.5], ParetoSPO shows a clear improvement in performance over Scalarized SPO and two stage MSE on both problems and across all noise settings.}
% \end{figure}

Our results indicate a clear performance hierarchy. The standard MSE baseline performs reasonably well but is consistently outperformed by the decision-focused methods. The scalarized SPO+ baseline shows a marked improvement over MSE, demonstrating the value of incorporating the optimization task into the learning objective.

Notably, our proposed \texttt{ParetoSPO} method consistently outperforms the three baselines in both the problem domains and all levels of noise. Even in the low-noise setting, standard SPO+ baseline, limited by linear scalarization, struggles to find solutions in these non-convex regions. In contrast, \texttt{ParetoSPO}, leveraging knowledge of the true Pareto Front to select its target, successfully guides the model towards these complex trade-offs, resulting in a significantly lower hypervolume regret. As noise increases, all methods degrade, but \texttt{ParetoSPO} maintains a clear advantage.

\section{Discussion, Limitations and Future Work}

The experimental results confirm that \texttt{ParetoSPO} effectively addresses the research gap identified at the intersection of DFL and multi-objective optimization. Existing methods have forced a choice between the computational tractability of scalarization-based DFL and the theoretical completeness of Pareto-based optimization. Our method provides a practical and effective way to achieve the best of both worlds.

The core innovation is the design of the \texttt{ParetoSPOLoss} function. By determining the target solution $w^*$ from the true Pareto Front, rather than from a potentially flawed scalarization of the true costs, we provide a much richer and more geometrically sound learning signal. This allows the model to learn predictions that induce plans in the correct region of the objective space, even if that region is non-convex and thus invisible to the standard SPO+ method. The superior performance of \texttt{ParetoSPO} directly quantifies the value of incorporating this explicit knowledge of the Pareto Front's geometry into the decision-focused learning process.

This work advances DFL by providing a practical methodology for building AI systems that can be used for decision-making and planning. It demonstrates how to retain the computational efficiency of single-objective DFL while reaping the theoretical benefits of a Pareto-aware approach. 
% This contribution can be seen not just as a solution to a specific problem, but as a blueprint for a new class of DFL loss functions where the intelligent selection of the target solution is a key design consideration.

While our proposed \texttt{ParetoSPO} method demonstrates strong empirical performance, it is important to acknowledge its limitations, which in turn motivate several promising directions for future research.

The primary limitation is the scalability of the ``model committee'' architecture. Our approach requires training a separate, specialized model for each known user preference. This means that training time scales linearly with the number of discrete preferences. While this process is highly parallelizable, it may become computationally prohibitive for problems with a very large or continuous spectrum of user preferences.

Furthermore, this architecture inherently lacks the ability to generalize to new, unseen preferences. Each expert model is trained exclusively on data corresponding to its single, fixed preference vector and has no knowledge of other trade-offs. Consequently, if a user presents a preference that was not part of the original training set, the system cannot generate a tailored, high-quality decision.

These limitations highlight a clear and compelling path for future work. The most significant next step is to overcome the scalability and generalization issues by developing a single, preference-conditioned model. Such a model would take both the feature vector $x$ and the preference vector $p$ as inputs, learning a continuous mapping from preferences to optimal decisions. This could potentially be achieved using techniques from meta-learning or by adapting scalable multi-objective learning architectures like COSMOS \citep{cosmos} to the DFL paradigm. Success in this area would represent a major advance towards building truly flexible and adaptive planning systems.

Additional future work includes applying \texttt{ParetoSPO} to more complex, real-world planning problems beyond the benchmarks used here, as well as integrating more sophisticated preference elicitation methods to handle scenarios where user preferences are not explicitly known beforehand.

\section{Conclusion}
In this paper, we introduced \texttt{ParetoSPO}, a novel decision-focused learning framework designed to address the unique challenges of multi-objective combinatorial optimization problems. By integrating a Pareto-informed target selection mechanism into the SPO+ loss function, our method successfully overcomes the established limitations of linear scalarization, leading to superior decision quality, particularly for problems with non-convex Pareto Frontiers.

Our experiments on the Multi-Objective Shortest Path and Multi-Objective Knapsack problems demonstrate that \texttt{ParetoSPO} consistently and significantly outperforms both traditional two-stage and standard scalarized DFL baselines across various noise levels. This work provides a practical and effective methodology for building AI systems that can learn to make high-quality, preference-aligned decisions under uncertainty. By showing how to bridge the gap between the computational efficiency of single-objective DFL and the theoretical completeness of Pareto-based methods, we offer a blueprint for a new class of DFL loss functions where the intelligent selection of the learning target is a key design consideration.

% Our experiments on the Multi-Objective Shortest Path and Knapsack problems demonstrate that \texttt{ParetoSPO} consistently outperforms standard two-stage and scalarized DFL baselines. This work provides a practical and effective methodology for building AI systems that can learn to make high-quality, preference-aligned decisions under uncertainty in complex, multi-objective domains.

% The acknowledged limitations of this work serve as direct bridges to future research. The ``model committee" architecture requires training a separate model for each known preference and cannot generalize to unseen preferences. Future work could explore the development of a single, preference-conditioned model that can generalize across a continuum of preferences, potentially leveraging techniques from meta-learning. Furthermore, extending this framework to handle even larger-scale problems and integrating more sophisticated preference elicitation methods are promising avenues for continued research in the quest to build truly intelligent and adaptive planning systems.

% Future Work:
% - Developing a single model that can handle a continuous range of preferences.
% - Exploring methods for generalizing to unseen preferences, perhaps using techniques from meta-learning or transfer learning.
% - Applying ParetoSPO to more complex, real-world planning problems.

%%%%%%%%%%%%%%%%%%%%%%%%%%%%%%%%%%%%%%%%%%%%%%%%%%%%%%%%%%%%%%%%%%%%%%%%

%%% Use this command to include your bibliography file.

\bibliography{./main}

\end{document}